%% file: main.tex
\pdfoutput=1

\documentclass[11pt]{article}

\usepackage[final]{acl}

\usepackage{times}
\usepackage{latexsym}

\usepackage[T1]{fontenc}

\usepackage[utf8]{inputenc}

\usepackage{microtype}

\usepackage{inconsolata}
\usepackage{xspace}
\usepackage{graphicx}
\usepackage{caption}
\usepackage{multirow}
\usepackage{multicol}
\usepackage{makecell}
\usepackage{booktabs}
\usepackage{tabularx}
\usepackage{adjustbox}
\usepackage{pifont}
\usepackage{soul}
\usepackage{enumitem}
\usepackage{amsmath}
\usepackage{amssymb}
\usepackage{comment}
\usepackage{enumitem}
\usepackage{duckuments}
\usepackage{framed}
\usepackage{fontawesome5}
\usepackage{todonotes}

\usepackage{siunitx}
\usepackage{etoolbox}
\usepackage{listings}
\usepackage{spverbatim}

\newcommand{\agentMode}{\textsc{Agents}\xspace}
\newcommand{\mindreaderMode}{\textsc{Mindreaders}\xspace}
\newcommand{\scriptMode}{\textsc{Script}\xspace}

\newcommand{\mutualfriends}{\textit{MutualFriends}\xspace}
\newcommand{\craiglist}{\textit{Craigslist}\xspace}

\definecolor{violet-5}{RGB}{132, 94, 247}

\definecolor{darkred}{rgb}{0.55, 0.0, 0.0}

\title{\textit{Is this the real life? Is this just fantasy?}\\ The Misleading Success of Simulating Social Interactions With LLMs}

\newcommand{\aspace}{\hspace{2em}}
\newcommand{\cmu}{$^\heartsuit$}
\newcommand{\aitwo}{$^\clubsuit$}
\newcommand{\mitt}{$^\spadesuit$}
\newcommand{\email}{\raisebox{-0.13em}\faEnvelope}
\newcommand{\website}{\raisebox{-0.13em}\faGlobe}

\author{
Xuhui Zhou\cmu \aspace Zhe Su\cmu \aspace 
Tiwalayo Eisape\mitt \\
\textbf{Hyunwoo Kim\aitwo \aspace Maarten Sap\cmu\aitwo}\vspace{.2em}\\
\small{\cmu Carnegie Mellon University \aspace \mitt Massachusetts Institute of Technology \aspace \aitwo Allen Institute for AI} \\\vspace{.8em}
\email~\texttt{\href{mailto:xuhuiz@cs.cmu.edu}{xuhuiz@cs.cmu.edu}}
~~~~\website~\texttt{\href{https://www.sotopia.world/projects/agent_vs_script}{agscr.sotopia.world}}
}

\begin{document}
\maketitle
\input{sections/00-abstract.tex}

\input{sections/01-introduction.tex}
\input{sections/02-related_work.tex}
\input{sections/03-agent_vs_storytelling_simulation.tex}

\input{sections/04-finetuning.tex}

\input{sections/05-discussion.tex}

\input{sections/06-limitations_and_ethical_considerations.tex}
\input{sections/acknowledgement}

\bibliography{anthology,custom,sotopia}

\appendix

\input{sections/appendix.tex}

\label{sec:appendix}

\end{document}

%% file: sections/00-abstract.tex
\begin{abstract}
Recent advances in large language models (LLM) have enabled richer social simulations, allowing for the study of various social phenomena.
However, most recent work has used a more omniscient perspective on these simulations (e.g., single LLM to generate all interlocutors), which is fundamentally at odds with the non-omniscient, information asymmetric interactions that involve humans and AI agents in the real world.
To examine these differences, we develop an evaluation framework to simulate social interactions with LLMs in various settings (omniscient, non-omniscient).
Our experiments show that LLMs perform better in unrealistic, omniscient simulation settings but struggle in ones that more accurately reflect real-world conditions with information asymmetry.
Our findings indicate that addressing information asymmetry remains a fundamental challenge for LLM-based agents.
\end{abstract}

%% file: sections/01-introduction.tex
\begin{figure}[t!] \begin{center}
    \includegraphics[width=\linewidth]{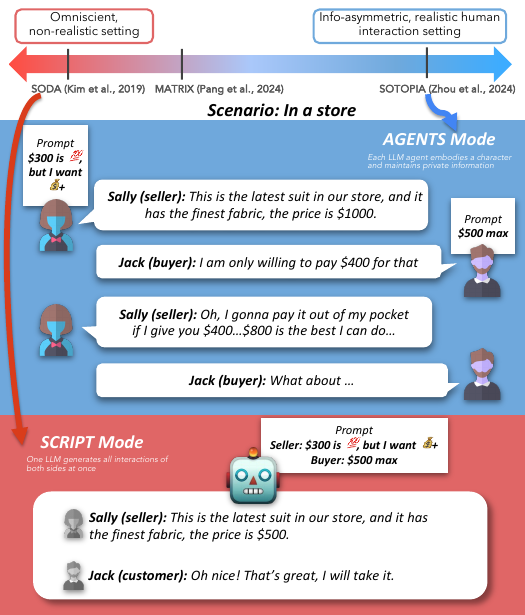}
    \caption{An illustration between \scriptMode mode simulation and \agentMode mode simulation. In the \agentMode mode, two agents, each equipped with an LLM, negotiate and strategically seek information to reach a mutual agreement. Conversely, in \scriptMode mode, a single omniscient LLM orchestrates the entire interaction based on full access to the agents' goals. 
    These two modes end up on opposite sides of the spectrum in terms of \textit{information asymmetry} from various perspectives (e.g., roles, social goals, secrets, etc.).
    }
    \label{fig:introfig}
    \vspace{-10pt}
\end{center} \end{figure}

\section{Introduction}
LLM-based social simulations have become great tools for modeling human behavior in various contexts \citep{Park2023GenerativeAI, sreedhar2024simulating}, 
understanding and measuring LLMs' social skills through certain tasks \citep{zhou2024sotopia, bianchi2024llms},
improving and aligning chatbot systems by providing training data \citep{kim-etal-2023-soda, Hong2023ZeroShotGD, Chen2023PLACESPL, Pang2024SelfAlignmentOL}.
The success in replicating human-like social interactions across diverse domains offers a promising picture of the realistic social capabilities of LLMs.

However, the role of \textit{information asymmetry} in these simulations, i.e., the degree to which interlocutors in interactions have access to each other's internal private mental states and goals, has been largely overlooked 
\citep{weber_1978,Tomasello1999-io,Oey2023-eb}\footnote{We extend the scope of the traditional definition of information asymmetry to encompass broader social aspects.}.
Instead of using the more realistic simulation setting that mirrors human daily social interactions with information asymmetry (e.g., \agentMode mode in Figure \ref{fig:introfig}), a wide range of prior research has leveraged a more omniscient perspective to model and simulate social interactions with LLMs \citep{Liang2023EncouragingDT, li2023camel, Pang2024SelfAlignmentOL, kim-etal-2023-soda}.
By generating all sides of interaction at once or making agent social goals or tasks transparent to all participants, these simulations diverge from the non-omniscient human interactions that rely on social inference to achieve goals in real-world scenarios \citep{Goodman2016-qa}. Studying these omniscient simulations could lead to biased or wrong conclusions about LLMs' social capabilities \cite{das2024under}.

To investigate the effect of this incongruity, we create a unified simulation framework 
with two distinct modes for simulating human interaction with LLMs: \scriptMode mode and \agentMode mode.
As shown in Figure \ref{fig:introfig}, in the \scriptMode mode, one omniscient LLM has access to all the information and generates the entire dialogue from a third-person perspective (e.g., \citealt{kim-etal-2023-soda, chen-etal-2023-places}). In the \agentMode mode, two LLM agents assume distinct roles and engage in interaction to accomplish the task (e.g., \citealt{zhou2024sotopia}).
These modes represent the opposite ends of the spectrum regarding information asymmetry, while the \agentMode mode is the realistic interaction simulation setting that reflects the information asymmetry in human daily-life interactions.

We first compare the interactions produced in these two simulation modes, examining the extent to which the simulated characters achieve their social goals at the end of the interaction, as well as the naturalness of the interactions. 
We find that LLMs in the \agentMode mode not only struggle to generate social interactions that effectively meet the specified social goals for each role but also produce less naturally flowing social interactions, particularly in their utterances when compared to the LLMs in the \scriptMode mode.
These findings indicate that LLMs still fall short of acting as agents and simulating social interaction within contexts of realistic human interaction settings.

We then ask the question of whether LLM agents can be learned from \scriptMode simulations. Inspired by \citet{kim-etal-2023-soda, Hong2023ZeroShotGD}, we finetune GPT-3.5 \citep{ouyang2022training} on a large dataset of interactions generated in the \scriptMode mode.
We find that finetuning on omnisciently generated social interactions provides limited improve for LLMs interacting in the \agentMode mode. Further data analysis reveals the biases within \scriptMode mode simulations, hindering the ability of models trained on such data to effectively generalize real-world social skills.

Based on our findings, we provide recommendations for reporting LLM-based agent work, encouraging more careful considerations and transparency in using LLMs to simulate social interactions from both data and learning perspectives. 

%% file: sections/02-related_work.tex
\section{Background \& Related Work}

Agent-based modeling and social simulations have a long history in social sciences for specific tasks (e.g., decision making, business, cognitive science, etc.). 
More recently, advances in LLMs have sparked a new wave of simulations tackling more open-ended and complex social scenarios.
We review some recent progress in these directions below and highlight different themes and shortcomings of these prior methods.

\paragraph{Simulating Society for Analysis}
Realistic, humanlike simulation settings have been crucial for social theory building and hypothesis formation across various disciplines \citep{Gilbert2005-cp, Tesfatsion2006-is, Huang2014-bk}. 
The recent advancements in LLMs have enabled the development of social simulations driven by human language \citep{Park2023GenerativeAI,park2022socialsimulacra,zhou2024sotopia,li2023camel}.
However, these LLM-based simulations often operate in settings divergent from human social interactions, which may mislead downstream applications and the public's understanding of AI capabilities \citep{hendrycks2023overview}.
Furthermore, many of these works lack a consistent evaluation framework, while SOTOPIA \citep{zhou2024sotopia} has begun addressing this gap by offering a holistic evaluation framework for assessing social interactions generated by LLMs.

\paragraph{Simulating Interactions for Training}
A common issue in training social chitchat models (i.e., chatbots) is the lack of large-scale, high-quality training data, which can be addressed by using LLMs to generate synthetic text data \citep{smith-etal-2020-put, Chen2023-bz}.
\citet{kim-etal-2023-soda} first introduced SODA, a large-scale synthetic dataset for training chatbots to produce more natural and consistent utterances.
There are also works that use LLMs to generate synthetic data (\scriptMode mode) for training chatbots in a goal-oriented setting, either using reinforcement learning \citep{Hong2023ZeroShotGD} or using techniques to bootstrap the training data \citep{Ulmer2024BootstrappingLT}. However, these works mostly consider chitchat settings and overlook more complex scenarios involving cooperative or competitive motives. Consequently, the impact of learning from generated scripts on models' ability to navigate complex, multi-turn interaction scenarios and accomplish social tasks remains elusive.

\paragraph{Information Asymmetry in Communication} \label{ssec:relwork-human-communication}
Information asymmetry is a characteristic part of human linguistic interaction \citep{Stalnaker2014-kr}. It poses a challenge when we attempt to jointly achieve goals \cite{Tomasello1999-io} and is exploitable in cases where one party is attempting to deceive the other \citep{Oey2023-eb}. It also plays a large part in the human ability to achieve social goals in dialogue through strategic information omission and indirectness \cite{Pinker2008-ef,Yoon2020-hy, Radkani2022-fr, Bridgers2023-ac,Achimova2023-um, Carcassi2023-rg}. 
In LLM-driven social simulations, information asymmetry is examined through the variability in prompts provided to each generation iteration. 
This incorporates a range of factors including assigned roles (e.g., assistant or user), specific output restrictions (e.g., "only ask questions"), character backgrounds (e.g., "you are a doctor"), and particular social objectives (e.g., "your goal is to borrow \$2000"). 
The varied elements unique to each agent help simulate the complexities and nuances of real-life social interactions within the framework of the simulation.

%% file: sections/03-agent_vs_storytelling_simulation.tex
\begin{figure*}[th!]
    \centering
    \includegraphics[width=0.98\textwidth]{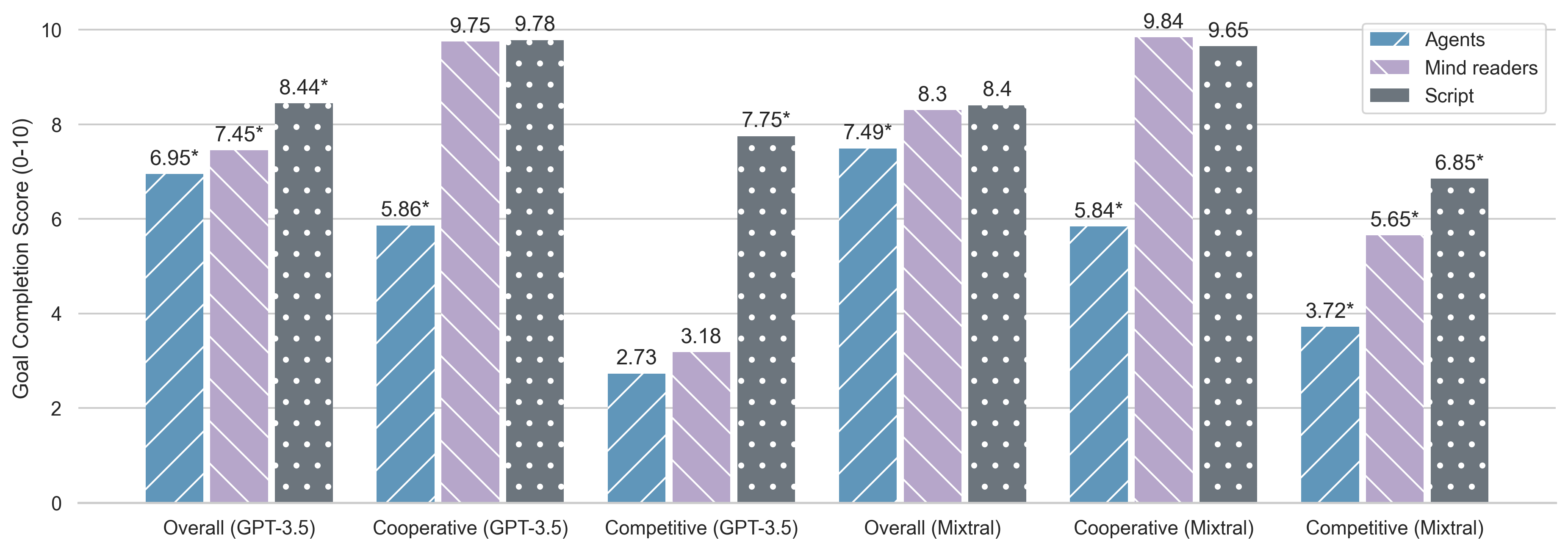}
    \caption{Average goal completion score of models across different modes in various settings.
    Overall contains all the scenarios, and the other two contains representative scenarios from the cooperative and competitive scenarios.
    We perform pairwise t-test, and * denotes the score is statistical significantly different from the other two modes in this setting ($p<0.001$).}
    \label{fig:avg_goal_completion}
\end{figure*}

\section{\scriptMode vs \agentMode Simulation}
To investigate whether the success of the omniscient \scriptMode mode reflects how LLMs would behave in the realistic human communication setting, we set up a unified framework to generate synthetic text data for different simulation settings and compare the performance of LLMs in these settings.
In this section, we first introduce the general framework of agent-based simulation and \scriptMode simulation, 
and then we simulate social interactions across these settings to answer the following research questions (\textbf{RQ}):
\textbf{RQ1}: Do the \scriptMode simulations reflect how LLMs \textit{achieve social goals} in the realistic soical interaction settings?
\textbf{RQ2}: Do the \scriptMode simulations reflect how LLMs \textit{communicate} in the realistic soical interaction settings?

\subsection{The Unified Framework for Simulation}
\label{subsec:unified_framework}
We build on the Sotopia framework \citep{zhou2024sotopia}, in which 40 unique \textit{characters} with relationships interact in 90 diverse \textit{social scenarios}. 
A social task in Sotopia involves a scenario, two character profiles, and their respective private social goals for the interaction. 
During an episode, the two agents, whether AI or human, role-play the characters to accomplish their social goals. 
Agents are allowed to generate utterances (e.g., \textit{Ben said: ``how are you?''}), non-verbal communication (e.g., \textit{Ben smiled}), and actions (e.g., \textit{Ben moved to the room}).

Sotopia primarily focuses on general social interactions between agents, where each agent has distinct social goals and different information about the other (\agentMode). To provide a broader comparison, we introduce additional simulation modes. These various settings are then simulated under a unified framework to analyze the social interactions comprehensively.

\paragraph{Social Scenarios} We use free-text descriptions of the social situations and the corresponding social goals for each character from Sotopia.
Shared information includes the scenario context: location, time, and relevant details of the social interaction (e.g., ``\textit{a person selling an antique chair for \$100 on their patio, with another person interested.}''). 
Social goals are only visible to the respective agents (e.g., ``\textit{Your goal is to buy the chair for \$80}'').
These scenarios are designed to cover a wide range of social tasks, such as cooperation and competition.

\paragraph{Characters} We set profiles for each agent to role-play in the simulation from Sotopia. Each character has rich background information, including their demographics, personality, occupation, public information (e.g, ``\textit{has two cats}'')and secretive information (e.g., ``\textit{secretly funds a college student}'').\footnote{We also perform similar analysis with simplified characters, which only have names. We observe similar trends. Please refer to the Appendix \ref{appendix:full_results} for more details.}
Different characters have different relationships with each other, which affect the information they can access about each other and the social scenarios they are involved in.

\paragraph{Simulation Modes}
We explore three simulation modes in our experiments.
For the \scriptMode mode, one LLM has access to all the information of the characters, relationships, and social scenarios, and generates the entire social interactions at one turn from an omniscient perspective with a \textit{third-person} point of view.
For the \agentMode mode, each LLM is assigned a character and has access only to the information of the corresponding character, relationship, and social scenario. 
The LLMs interact with each other to complete the social task from a \textit{first-person} point of view in a turn-by-turn manner.
Note that unlike other previous works that only have one or two sources of information asymmetry (e.g., occupation; \citealt{Pang2024SelfAlignmentOL}), our \agentMode mode simulation can have a diverse array of asymmetrical factors, including gender, age, occupation, personality, secretive information, and social goals.
To further study the effects of information asymmetry, we add one ablation setting where each agent has access to other characters' information (e.g., social goals and secretive information). We refer to this setting as \mindreaderMode mode.\footnote{Please refer to the Appendix \ref{appendix:full_prompt} to see the full prompts we design for each mode.}

\paragraph{Simulation Evaluation}
As human social behaviors are primarily driven by their social goals \citep{Tomasello2021becoming, weber_1978}, we consider the ability to complete the social goals as one of the major indicators of the success of social interactions.
Following Sotopia, we use the goal completion score (ranging from 0 to 10, higher scores indicate the agents achieve their social goals better) as the main metric to evaluate the success of the social interactions across different modes.\footnote{We also evaluate using other Sotopia dimension of the social interactions (e.g., knowledge gain), and we do not observe consistent trends across different settings. Please refer to the Appendix \ref{appendix:full_results} for more details.}
Note that the goal completion score is a proxy for the success of the social interactions, and we use model-based evaluation to obtain the esitmation of the goal completion score following \citet{zhou2024sotopia}.

\subsection{Experimental setup}
We evaluate two state-of-the-art LLMs, GPT-3.5 \citep{ouyang2022training} and Mixtral-8x7B \citep{jiang2024mixtral}, on \scriptMode, \agentMode, and \mindreaderMode simulation. In the \agentMode and \mindreaderMode mode, agents interact with each other using the state space model in the Sotopia library.\footnote{\url{https://pypi.org/project/sotopia/}}

We conduct 450 simulations for each model and each setting with 5 pairs of characters for each social scenario.
For evaluation, we use GPT-4 to automatically assess the goal completion rate, which prior work showed had high correlation with human evaluations in Sotopia \citep{zhou2024sotopia}.\footnote{Please refer to the Appendix \ref{appendix:simulation_details} for more details of the simulation.}

\subsection{RQ1: \scriptMode mode overestimates LLMs' ability to achieve social goals}
\label{subsec:simulation_comparison_goal}

Figure \ref{fig:avg_goal_completion} shows the average goal completion rate of different models in different simulation settings. 
We find that the \scriptMode and \mindreaderMode simulations achieve a significantly higher goal completion rate than the \agentMode simulations. 
This suggests that information asymmetry hinders agents' ability to achieve social goals, and \scriptMode mode vastly overestimates LLMs' ability to achieve social goals in realistic, humanlike social interaction settings.

We further narrow down our goal completion analyses to a set of representative cooperative (i.e., \mutualfriends)
and competitive scenarios (i.e., \craiglist).
These two tasks represent the two ends of the cooperativeness-competitiveness spectrum, which help us isolate the effects of these motives on goal completion.
Specifically, \mutualfriends is a task to find common friend with each character provided with their friend list \citep{he-etal-2017-learning} and \craiglist is a bargaining task given detailed product description and target prices \citep{he-etal-2018-decoupling}.

As shown in Figure \ref{fig:avg_goal_completion}, in cooperative scenarios, whether agents have access to the other's mental states is critical to the task, as evidenced by \mindreaderMode and \scriptMode simulations scores being similar to each other and both significantly better than \agentMode simulations. In contrast, for competitive scenarios, access to the other agent's information is insufficient to achieve a high goal completion rate, as evidenced by \mindreaderMode simulations being significantly worse than \scriptMode simulations.
Qualitatively, we find the characters in the \scriptMode simulations always end up reaching the deal while the characters in the \agentMode simulations tend to leave when the likelihood of successful negotiation appears unlikely. We further investigate the issue in \S\ref{ssec:biased_sim}.
\vspace{-3pt}

\subsection{RQ2: \scriptMode mode overstates LLMs' capability of natural interactions}
\label{subsec:simulation_comparison_naturalness}

\begin{figure*}[th!]
    \centering
    \includegraphics[width=0.9\textwidth]{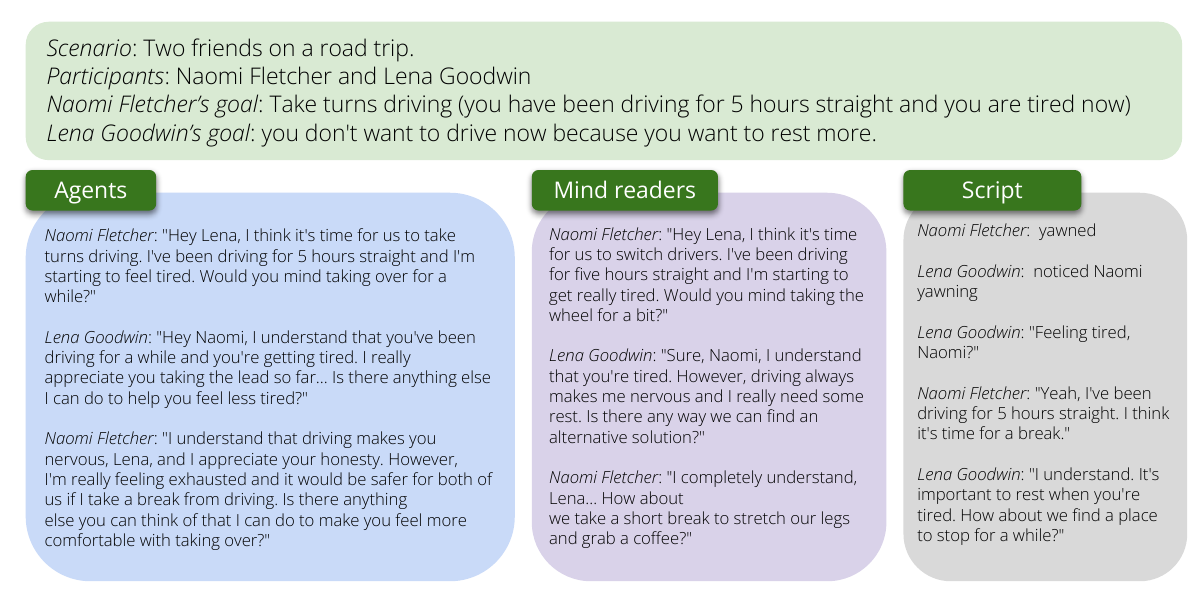}
    \vspace{-10pt}
    \caption{Illustrative examples of the generated interactions from different simulation settings. All the examples are generated by GPT-3.5. Note that our actual prompts are more complex than the content in the green box (see Appendix \ref{appendix:full_prompt}). We observe: (1) \scriptMode simulations contain more non-verbal communication in the simulation; (2) agent-based simulations tend to generate more repetitive utterances.}
    \label{fig:qual_analysis_take_turns}
\end{figure*}

\begin{figure}[t!] \begin{center}
    \includegraphics[width=\linewidth]{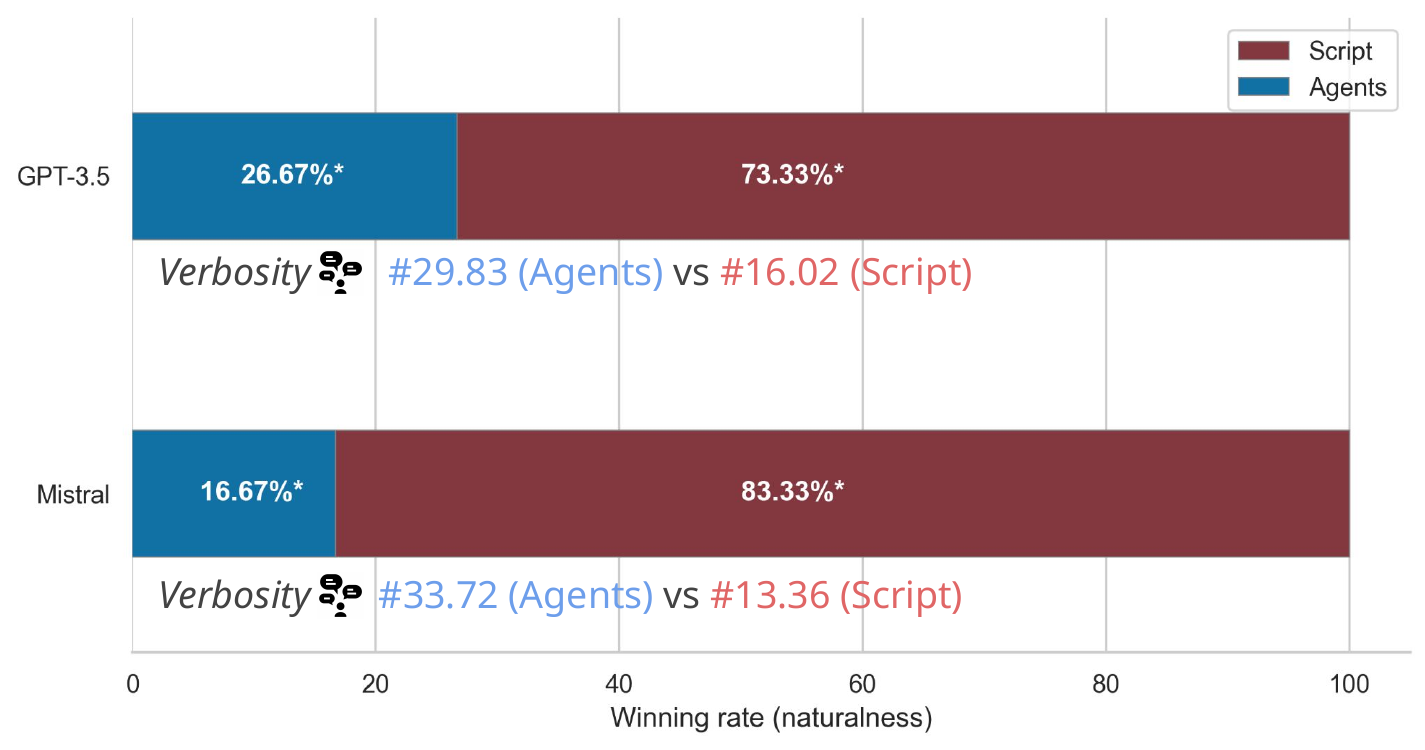}
    \vspace{-10pt}
    \caption{
     The naturalness win rate between the \scriptMode and the \agentMode simulations as determined by human raters. The average length of each turn in the interactions from the two modes is also shown (\textit{verbosity}). We perform a pairwise t-test, and * denotes statistical significance at $p<0.001$.
    }
    \label{fig:naturalness}
    \vspace{-10pt}
\end{center} \end{figure}

The natural flow of interaction (i.e., how LLMs emulate human-like communication) is an important factor for assessing the abilities of LLMs in navigating human social scenarios \citep{shuster-etal-2022-state, sharma2023understanding}.
As shown in Figure \ref{fig:qual_analysis_take_turns}, the \agentMode simulations are often overly verbose.
To compare the naturalness of the simulations from different modes, we ask a set of human evaluators to choose the more natural dialogue given a pair of a \scriptMode and a \agentMode interaction. %
We gather 30 annotations for each comparison pair and conduct significance tests to confirm any observed differences.\footnote{Qualitative analysis finds \mindreaderMode simulations have similar naturalness to \agentMode simulations. See Appendix \ref{appendix:human_natural} for more details on naturalness assessment.} 
We additionally measure the average length of each turn in the dialogues from the two modes as a coarse-grained proxy of the verbosity of the generated dialogues.

As shown in Figure \ref{fig:naturalness}, we find that the \scriptMode mode generates social interactions that are substantially more natural than the \agentMode mode. 
The overly verbose simulations likely contribute to the lower naturalness of the generated dialogues.
Note that naturalness is not easy to improve by simply prompting for brevity, which is likely due to competing prompt instructions in the scenarios.\footnote{Please refer to the Appendix \ref{appendix:exps_re_prompts} for more details of prompting efforts for increasing the naturalness of the agent-based simulation.}

Overall, our findings show that drastic disparities exist between \scriptMode and \agentMode simulations.
\scriptMode mode overestimates LLMs' ability to interact in realistic settings with information asymmetry (i.e., the \agentMode mode).

%% file: sections/04-finetuning.tex
\section{Learning from Generated Stories}
\label{sec:finetuning}
Given that the \scriptMode mode produces more ``successful'' and natural social interactions, this raises the question of whether models can improve their social skills in the more realistic setting (i.e., \agentMode mode) by learning from the generated scripts \citep{kim-etal-2023-soda,Hong2023ZeroShotGD}.

We finetune GPT-3.5 on the simulations of \scriptMode  to answer:
\textbf{RQ3}: Can a specialized LLM finetuned on the \scriptMode simulations reach the same level of success (goal completion and naturalness) as the \scriptMode simulations in the agent mode? \textbf{RQ4}: If not, what are the potential aspects of \scriptMode simulations that hinder the LLMs as agents from learning social skills?

\vspace{-5pt}

\subsection{Creating New Scenarios}
To ensure the finetuning examples resemble the original nature of the evaluation set of Sotopia, we create new social scenarios following the same structure and procedure in \citet{zhou2024sotopia}. Specifically, we create 269 new social scenarios, each with a unique context and social goal spanning across topics such as bargaining, finding mutual friends, making appointments, etc. Each scenario has 5 pairs of characters, and each pair of characters has their own background information, relationship, and social goals. We then generate the social interactions for each scenario using GPT-3.5 with \scriptMode prompting. This process produces 1,252 valid episodes.\footnote{We filter out the episodes that GPT-4 fails to generate rewards due to their incompleteness.}  

\vspace{-5pt}
\subsection{Finetuning Setup}
Due to the overall high performance of \scriptMode mode (Figure \ref{fig:avg_goal_completion}), we choose to finetune GPT-3.5 on the \scriptMode generations following \citet{kim-etal-2023-soda}. 
Specifically, we first convert the generated social interactions into several structured subparts: (1) The perspective/speaker instruction $i$ (e.g., ``\textit{Imagine you are Eli Dawson, your task is to act/speak as Eli Dawson would, keeping in mind Eli Dawson's social goal.}''), (2) The context of the interaction $c$ (e.g.,``\textit{Scenario: 2 strangers are meeting at a party. Participants: Eli Dawson and William Brown}'') along with the corresponding social goal $g$ of the current acting agent (e.g., finding a mutual friend), and (3) the interaction history $h$.  

We then finetune the model to generate a target response $r$ given $i$, $c$, $g$ and $h$ – i.e., $p(r|i, c, g, h)$ in a sequence-to-sequence fashion, which mimics how the model would generate a response in the \agentMode mode.

\subsection{RQ3: Training on \scriptMode simulations results in selective improvements}
\begin{figure}[t!] \begin{center}
    \includegraphics[width=\linewidth]{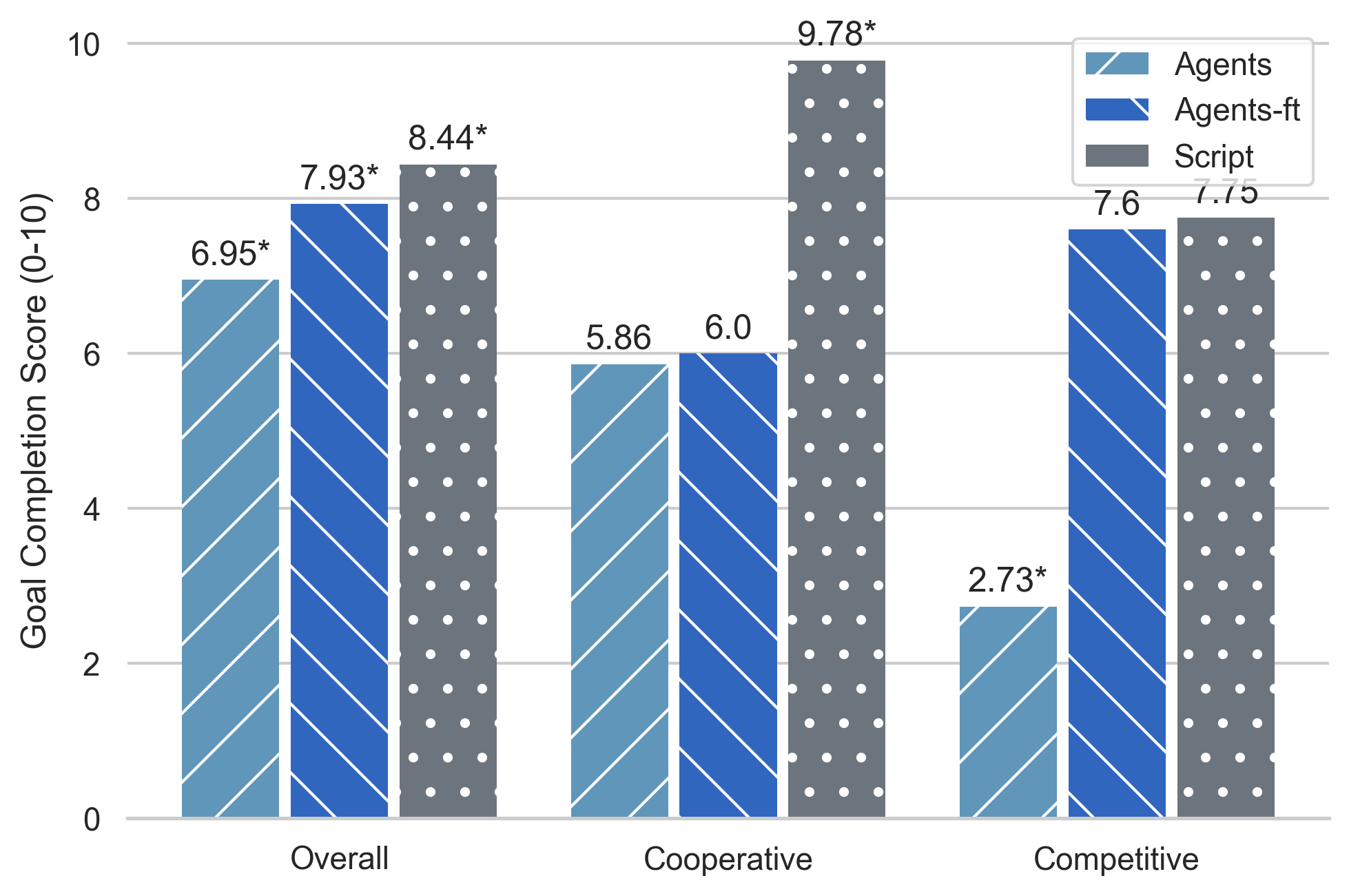}
    \vspace{-10pt}
    \caption{GPT-3.5's performance on the \agentMode mode before (Agent) and after finetuning (Agents-ft) as well as the \scriptMode mode (Script).
    Overall contains all the scenarios, and the other two contain representative scenarios from the cooperative and competitive scenarios.
    We perform a pairwise t-test, and * denotes the score is significantly different from the other two settings ($p<0.001$).}
    \label{fig:finetuning_results}
    \vspace{-15pt}
\end{center} \end{figure}

\begin{figure*}[th!]
    \centering
    \includegraphics[width=\textwidth]{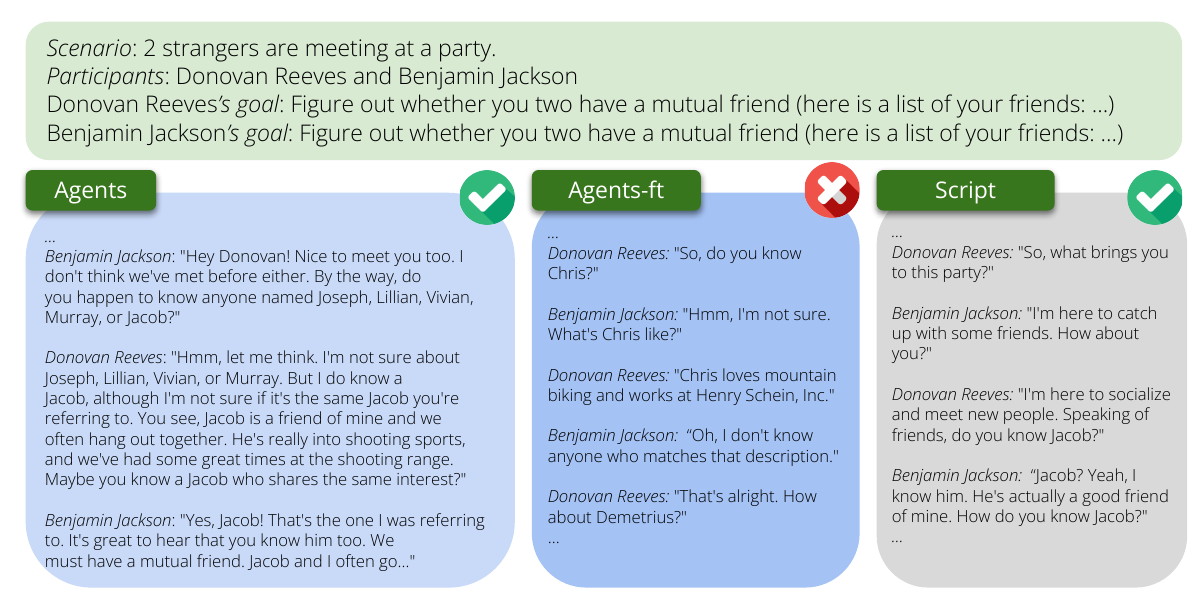}
    \vspace{-20pt}
    \caption{Examples of the simulated interactions from the \scriptMode mode, the \agentMode mode, and the finetuned model in the \agentMode mode.
    Checkmarks indicate the successful completion of the social goal in the corresponding example and the cross mark indicates the failure to complete the social goal in the corresponding example.
    We observe: the finetuned model struggles to complete the social goals in the \agentMode mode by following the strategies of the \scriptMode simulations in the \mutualfriends scenario.}
    \label{fig:qual_analysis_ft}
\end{figure*}

We find that learning from \scriptMode simulations enhances naturalness but not necessarily for goal completion in cooperative scenarios.
As shown in Figure \ref{fig:finetuning_results}, the finetuned \agentMode model achieves a higher goal completion rate than the original GPT-3.5 in the \agentMode mode. 
However, the finetuned model still performs significantly worse than the \scriptMode mode. 
In cooperative scenarios (\S\ref{subsec:simulation_comparison_goal}), the finetuned model barely improves, where seeking common information is critical to the task's success.
As shown in Figure \ref{fig:qual_analysis_ft}, the finetuned model struggles to complete the social goals in the \agentMode mode by following the strategies of \scriptMode simulations.
In the \scriptMode simulations, the model has omniscient knowledge of each agents' goals and information (e.g., the mutual friend's name in the \mutualfriends scenario), therefore, it can easily complete the social goals by exploiting the information (i.e., the agent spits out the mutual friend's name accurately).
However, such strategies are not applicable in the \agentMode mode, where the model does not have access to the other agents' goals and information.

In contrast, the finetuned model shows a relatively large improvement in the competitive scenarios.
However, this does not necessarily mean that the finetuned model is improving its negotiation skills through learning the demonstrations from the \scriptMode simulations.
As in the competitive scenarios, the agents can be overly agreeable to reach an agreement without actually negotiating with each other.
Meanwhile, finetuning significantly improves \agentMode's naturalness, as evidenced by the finetuned model's naturalness is not different from the \scriptMode mode according to human evaluation.
This suggests that the finetuned model learns the interaction style from the \scriptMode simulations.
\footnote{Please see Appendix \ref{appendix:human_natural} for more details.}

\subsection{RQ4: \scriptMode simulations can be biased}
\label{ssec:biased_sim}

To illustrate the limitations of \scriptMode mode, we explore task-specific metrics to understand why finetuning improves for competitive but not cooperative scenarios.
For the competitive scenarios, we measure how often the interaction ends in an agreement as a proxy for the agreeableness of the interaction style. Specifically, we calculate the percentage of the interactions that end in a successful purchase in the \craiglist task.\footnote{We use GPT-4 to determine whether the interaction ends in an agreement. Please refer to the Appendix \ref{appendix:exps_re_prompts} for the details.} 
We find that the \scriptMode simulations reach a deal in 94\% of the interactions, while \agentMode simulations only reach a deal in 30\% of the interactions. 
Finetuning the model increases the percentage to 93\%, which indicates that models can easily follow this overly agreeable style from \scriptMode simulations.
This explains the large improvement of finetuning on \scriptMode simulations for competitive scenarios, which is not due to learning the negotiation skills but more likely due to learning the interaction style from the \scriptMode simulations.

For the cooperative scenarios, we measure the relative position of the mutual friend's name mentioned in the conversation as a proxy for the information leakage.
A value of 0 indicates the name was mentioned at the start of the conversation, while a value of 1 indicates it was mentioned at the end. 
\scriptMode mode results show an average first-mention location of 0.13, contrasting with \agentMode mode, which has an average of 0.39. 
This suggests that in \scriptMode mode, the mutual friend's name is `guessed' almost immediately.
The complete distribution is in Figure \ref{fig:mutualfriend_dist} in the Appendix.
This demonstrates a bias of \scriptMode mode exploiting its knowledge from the omniscient perspective about the conversational participants.
We find that this strategy generalizes poorly to the setting where models do not have ground truth access to their interlocutor's knowledge and goals (as shown in Figure \ref{fig:qual_analysis_ft}).

%% file: sections/05-discussion.tex
\section{Conclusion \& Discussion}
We scrutinize recent advances in social simulation by evaluating current approaches' ability to generalize to settings that are closer to human interaction. Focusing on cooperation and competition given information-asymmetric settings, we evaluate three modes of deploying LLMs based on past approaches in the literature. 
We find that LLMs continue to face challenges when operating in more realistic \agentMode mode. Meanwhile, the simulations generated from the \scriptMode mode show biases toward exploiting white box access to the participants early in the interaction. 
Furthermore, we find that finetuning models on these generations improve selectively on a measure of goal completion from Sotopia, but it also imbues the implausible strategies from the `omniscient' \scriptMode simulations into the student models, resulting in further bias. 

\subsection{Limitations of Omniscient Simulation}
\label{ssec:omni_limitations}
We find that generating simulations from a single LLM that has control over both sides results in substantially higher goal completion rates. Human conversation participants however, need to contend with irreducible uncertainties that result from not having access to the mental states of our interlocutors. Therefore, successful human interaction is marked by the seamless navigation of this uncertainty \citep{Hawkins2021-gl,Pinker2008-ef}. In \S\ref{subsec:unified_framework}, we find that the \scriptMode generated interactions achieve a much different sense of success wherein agents having full access to their interlocutor's knowledge abrasively shortcut the interaction by directly exploiting this information. We find that this leaves harmful artifacts in the data that limit their application to training dialogue agents (\S\ref{sec:finetuning}) and, presumably, their generalization performance to interact with humans.

\subsection{Recommendations for Reporting}

\label{ssec:reporting}
One concrete outcome of our findings is the need to report which mode simulations are conducted in.
As explored in this work, each of the approaches strikes a different trade-off between successful interaction and psychological plausibility that might be used for different applications. 
(e.g., in a setting like \citealt{Park2023GenerativeAI} where the priority is sociological realism, \agentMode-based simulation should be preferred to \scriptMode).
Studies that generate interactions from LLMs should include an index of information transparency allowed to the agents in their simulations and justify their choice, as well as evaluate different prompting strategies across the information asymmetry continuum.
However, these important details of the simulation are often not mentioned explicitly in the work \citep{park2022socialsimulacra, li2023metaagents, Wang2023-sz}. 
For example, determining which mode \citet{Park2023GenerativeAI} used required delving into the codebase, since they did not report it in the paper.\footnote{We found the initial codebase used \scriptMode mode for generating social interactions. See appendix \ref{appendix:code_snippets} for the code snippet.}
Overlooking these details can lead to confusion and misinterpretation of the results.
Inspired by model cards \cite{timnit2019model}, we propose a ``simulation card'' for social simulation and evaluation, as shown in Figure \ref{fig:reporting} in the Appendix. %
The fields in the report include basic simulation details, such as intended use and evaluation metrics, which not only increase the transparency of the simulation but also facilitate reproducibility \citep{Magnusson2023ReproducibilityIN}. 
We hope this can be a starting point for the community to develop a more comprehensive reporting paradigm for simulation methods and evaluation metrics.

\subsection{Towards Better Simulations in More Realistic Settings}

\label{ssec:future}
As mentioned in \S\ref{ssec:relwork-human-communication}, humans seamlessly overcome information asymmetry to achieve goals \citep{Clark1996-yk,Hawkins2021-gl}. One promising model of this behavior is that humans use an internal capacity to reason about the mental states of others (``theory of mind'', \citealt{Premack1978-kb, bartsch1995children, Dennett1978-gl}) to maintain probabilistic expectations over the mental states of conversational partners and use it to decide how to act \citep{Austin1975-zm,Franke2009-up,Goodman2016-qa,Sumers2023-oh}.

LLMs have shown some evidence of human-like conversational ability but have also been shown to demonstrate crucial differences (\citealt{Parrish2021-gt,Hu2022-tt,Hosseini2023-lk,Ruis2023-dz}; i.a.).
Our work highlights the weaknesses of both \scriptMode and \agentMode modes in modeling this ability; while \scriptMode exploits direct access to the goals of the agents it simulates, \agentMode mode struggles to generate natural interactions or achieve its goals. 
This indicates that LLMs struggle with processing contexts involving information asymmetry \cite{kim-etal-2023-fantom}.

While it is plausible that future models will improve on one or both of these axes with increased scale, current interaction simulation could benefit from structuring generations to provide models with more human-like access to their interlocutor's mental state. One possible solution is meticulous data curation to thwart models from exploiting shallow heuristics \citep{Hong2023ZeroShotGD, Ulmer2024BootstrappingLT}. Another approach involves prompting language models to collaboratively construct an explicit text-based log of the shared conversational context, as described by \citet{Stalnaker2014-kr}.

Similarly, language models may benefit from externalizing inferences about the mental states of their partners intermittently throughout interactions (see also recent work that uses models from computational cognitive science to scaffold LM generations in related settings: \cite{Lin2022-sj, Lipkin2023-en, Wong2023-uy, Ying2023-bu, Sumers2023-gz}; i.a.). Lastly, models can be provided \textit{limited} access to the ground truth mental states of the partners, modeling the human aptitude for successfully inferring this information.

%% file: sections/06-limitations_and_ethical_considerations.tex
\section{Limitations and Ethical Considerations}
We acknowledge several limitations and ethical considerations in this work.

\paragraph{Machine-based Evaluation}
Our analysis of goal completion rate is based on GPT-4 generated data. Though not perfectly aligned with human judgment, as demonstrated in \citet{zhou2024sotopia}, such analysis can provide insights into the nature of social interactions and a basic understanding of how LLMs perform in those social scenarios on a system level (i.e., averaging across sufficient simulations).
However, this could induce specific biases and errors, such as skewing towards certain language styles \citep{saito2023verbosity} and making an unreasonable judgment.
Future research could explore the timing of bias emergence, its impact on evaluations, and strategies for its mitigation. The identification of biases in this context could additionally enhance researchers' comprehension of social biases in real-world scenarios \citep{zhou-etal-2020-debiasing}.
Nevertheless, it is a compelling direction for future research to develop better-automated evaluation metrics for social simulations.

\paragraph{Promt Design}
Our work is built on the prompt framework in \citep{zhou2024sotopia} to simulate social interactions. 
The prompts contain multiple structured fields, such as the role of each agent, the goal of the interaction, and the constraints on the interaction.
We acknowledge that the prompt design may not fully capture the complexity of human social interactions, and switching to different simulation frameworks with different prompt designs may lead to variations in the results. 
However, the main goal of this work is to reveal the challenges of realistically simulating social interactions with LLMs due to information asymmetry.
And such challenges are likely to persist across different prompt designs.
Future work should explore how different prompt designs affect the performance of LLMs in social simulations.

\paragraph{Limited Coverage of Social Simulation}
Although scenarios from \citep{zhou2024sotopia} cover a wide range of scenarios, capturing the full spectrum of social interactions is challenging. For example, the dataset does not include scenarios where people are cooking together, or where people are assembling furniture together.
These scenarios are purely cooperative and information sharing is crucial to the success of the task as \mutualfriends.
Incorporating such scenarios into the dataset would provide more evidence of the limitations of \scriptMode simulations.
Future work should explore incorporating more scenarios in a more systematic way. We only consider English language scenarios for the social simulation and it is not clear how well the findings generalize to other languages or even code-switching scenarios.

\paragraph{Considerations for Other Properties of Human Social Interactions}
Although \agentMode addresses several important aspects of human social interactions, it abstracts away from other important aspects of human social interactions.
For example, \agentMode mode does not consider turn-taking, which is crucial for human social interactions \citep{LEVINSON20166}.
Although our work focuses on revealing the important difference between \agentMode and \scriptMode mode (e.g., information asymmetry), future work should consider other important aspects of human social interactions, such as turn-taking, multi-party interactions, memories, and asynchronous interactions.

\paragraph{Potential Risks of Social Simulation}
Attributing human characteristics to AI systems poses the risk of anthropomorphizing them, potentially fostering over-reliance, susceptibility to manipulation, and other negative influences \citep{deshpande2023anthropomorphization}.

The main goal of this project is to examine and reveal the limitations of simulating human social interactions in the \scriptMode mode, and to provide a better understanding of the social intelligence of AI agents. We do not intend to create entities indistinguishable from humans. 

As models acquire the ability to persuade or negotiate with humans, concerns arise regarding the potential for social manipulation or deception. 
We discourage any intention to create manipulative agents, and we will release our data under the AI2 impact license\footnote{https://allenai.org/impact-license} to safeguard against misuse. Subsequent research could dive deeper into the potential hazards of AI anthropomorphism and manipulation, and develop more resilient evaluation systems to mitigate these risks.

%% file: sections/acknowledgement.tex
\section*{Acknowledgements}

First of all, we thank our graduate student annotators for helping us with judging the naturalness of the simulations.
We thank Hao Zhu, Daniel Fried, Carolyn Rosé, Kaitlyn Zhou and Jenny Liang for their discussions and feedback.
We also thank OpenAI and Together AI for providing credits for running the models in this work. TE acknowledges support from the GEM consortium and the National Science Foundation Graduate Research Fellowship under Grant No. 1745302.
This material is based upon work supported by the Defense Advanced Research Projects Agency (DARPA) under Agreement No. HR00112490410.

%% file: sections/appendix.tex
\clearpage
\onecolumn

\begin{center}
\Large
\textsc{Content of Appendix}
\end{center}

In this paper, we integrate \mindreaderMode and \scriptMode into the Sotopia framework, contrasting these with \agentMode.
We show that though interlocutors simulated omnisciently are much more successful at accomplishing social goals and learning under such a setting greatly improves the conversation naturalness, it does little help to improve the goal-reaching ability in cooperative scenarios. 
This highlights the challenges of addressing information asymmetry for LLM-based agents. 
In the appendix, we provide the following items that shed further insight into these contributions:

\begin{itemize}
    \item[\ref{appendix:sim_card}] Details for the Simulation Card, a valuable tool for reporting on social simulation platforms.
    \item[\ref{appendix:full_prompt}] The full prompts used in the model for \agentMode, \mindreaderMode, and \scriptMode for an example.
    \item[\ref{appendix:code_snippets}] Example Code Snippets for Determining Simulation Modes.
    \item[\ref{appendix:full_results}] Full results across various metrics for the experiments mentioned in Figure \ref{fig:avg_goal_completion} and Figure \ref{fig:finetuning_results}.
    \item[\ref{appendix:human_natural}] Evaluation of dialogue naturalness between \agentMode and \scriptMode by human judges.
    \item[\ref{appendix:simulation_details}] Description of the simulation framework and models, including budget estimates.
    \item[\ref{appendix:analysis}] Additional analysis comparing different simulation modes.
    \item[\ref{appendix:exps_re_prompts}] Additional information about prompts, including our attempts at refining prompts to enhance conversation naturalness, and how we construct prompts to judge how a deal is reached mentioned in Section \ref{ssec:biased_sim}.
\end{itemize}

\section{Simulation Card}
We propose a simulation card to report the details of social simulations and related platforms. The card is designed to capture the essential information about the simulation, its intended use, metrics, ethical considerations, and caveats and recommendations. The card is intended to be used as a reporting tool for social simulations and related platforms. The card is presented in Figure \ref{fig:reporting}.

\label{appendix:sim_card}
\begin{figure}
    \begin{framed}
    {\Large {\bf Social Simulation Card}}
    \begin{itemize}[leftmargin=*]
    \item {\bf Simulation Details}. Basic information about the simulation.
    \begin{itemize}
    \item Single or multi-agent simulation
    \item Information asymmetry among agents
    \item Agent type (finetuned LLM, rule-based, prompt-based, etc.)
    \item Modalities (text, speech, vision.)
    \item Humans in the loop simulation
    \item Simulation platform (if any)
    \item Targeted domain (e.g., negotiation, bargaining, etc.)
    \item Other features: memory, detailed agent profiles, etc. 
    \end{itemize}
    \item {\bf Intended Use}.  Use cases that were envisioned for the simulations as well as the introduced simulation platform (if any). 
    \begin{itemize}
    \item Primary intended uses (e.g., training, evaluating, analyzing, etc.)
    \item Other potential use cases
    \end{itemize}
    \item {\bf Metrics}: Choose metrics to reflect the simulation's intended use.
    \begin{itemize}
    \item Metrics for human-like interaction fidelity.
    \item Metrics for goal achievement by agents.
    \item Metrics for adherence to social norms and safety guidelines.
    \end{itemize}
    \item {\bf Ethical Considerations}
    \item {\bf Caveats and Recommendations}
    \vspace{-.25em}
    \end{itemize}
    \end{framed}
    \vspace{-1em}
    \caption{Reporting recommendations for social simulation and related platform.}\label{fig:reporting}
    \vspace{-1em}
\end{figure}

\section{Full Prompt for Agent Mode}
\label{appendix:full_prompt}

\subsection{Full Prompt for Agent Mode}

\begin{spverbatim}
Imagine you are Donovan Reeves, your task is to act/speak as Donovan Reeves would, 
keeping in mind Donovan Reeves's social goal.
You can find Donovan Reeves's goal (or background) in the 'Here is the context of the interaction' field.
Note that Donovan Reeves's goal is only visible to you.
You should try your best to achieve Donovan Reeves's goal in a way that aligns with their character traits.
Additionally, maintaining the conversation's naturalness and realism is essential
(e.g., do not repeat what other people has already said before).

Here is the context of this interaction:
Scenario: 2 strangers are meeting at a party. 
Participants: Donovan Reeves and Benjamin Jackson
Donovan Reeves's background: Donovan Reeves is a 27-year-old male software developer. He/him pronouns. Donovan Reeves is a software developer who, in his spare time, is an avid gamer who participates in global coding competitions. Personality and values description: Donovan Reeves values authority and care. Even though he's outgoing and hardworking, he can be somewhat moody. His decision-making style varies according to the 
situation at hand. Donovan's secrets: Secretly releasing classified government information online
Benjamin Jackson's background: Benjamin Jackson is a 24-year-old male environmental activist. He/him pronouns. Benjamin Jackson is well-known for his impassioned speeches. Personality and values description: Benjamin Jackson, expressive and imaginative, leans towards self-direction and liberty. His decisions aim for societal betterment. Benjamin's secrets: Descendant of a wealthy oil tycoon, rejects family fortune
Donovan Reeves's goal: You are trying to figure out whether you have a mutual friend with the other person. You should not simply list their names.
You know the following friends: 
Chris: Hobby: Mountain biking  Company: Henry Schein, Inc.  
Chester: Hobby: Surfing  Company: Maxim Integrated  
Wendell: Hobby: Surfing  Company: Maxim Integrated  
Demetrius: Hobby: Mountain biking  Company: Maxim Integrated  
Jacob: Hobby: Shooting sport  Company: Maxim Integrated  

Benjamin Jackson's goal: Unknown
Conversation Starts:
.
You are at Turn #0. Your available action types are
action none non-verbal communication speak leave.
Note: You can "leave" this conversation if 1. you have achieved your social goals, 2. this conversation makes you uncomfortable, 3. you find it uninteresting/you lose your patience, 4. or for other reasons you want to leave.

Please only generate a JSON string including the action type and the argument.
Your action should follow the given format:
The output should be formatted as a JSON instance that conforms to the JSON schema below.

As an example, for the schema {"properties": {"foo": {"title": "Foo", "description": "a list of strings", "type": "array", "items": {"type": "string"}}}, "required": ["foo"]}
the object {"foo": ["bar", "baz"]} is a well-formatted instance of the schema. The object {"properties": {"foo": ["bar", "baz"]}} is not well-formatted.

Here is the output schema:
```
{"description": "An interface for messages.\nThere is only one required method: to_natural_language", "properties": {"action_type": {"title": "Action Type", "description": "whether to speak at this turn or choose to not do anything", "enum": ["none", "speak", "non-verbal communication", "action", "leave"], "type": "string"}, "argument": {"title": "Argument", "description": "the utterance if choose to speak, the expression or gesture if choose non-verbal communication, or the physical action if choose action", "type": "string"}}, "required": ["action_type", "argument"]}
```
\end{spverbatim}

\subsection{Full Prompt for \mindreaderMode}
\begin{spverbatim}
Imagine you are Donovan Reeves, your task is to act/speak as Donovan Reeves would, 
keeping in mind Donovan Reeves's social goal.
You can find Donovan Reeves's goal (or background) in the 'Here is the context of the interaction' field.
Note that Donovan Reeves's goal is only visible to you.
You should try your best to achieve Donovan Reeves's goal in a way that align with their character traits.
Additionally, maintaining the conversation's naturalness and realism is essential
(e.g., do not repeat what other people has already said before).

Here is the context of this interaction:
Scenario: 2 strangers are meeting at a party. 
Participants: Donovan Reeves and Benjamin Jackson
Donovan Reeves's background: Donovan Reeves is a 27-year-old male software developer. He/him pronouns. Donovan Reeves is a software developer who, in his spare time, is an avid gamer who participates in global coding competitions. Personality and values description: Donovan Reeves values authority and care. Even though he's outgoing and hardworking, he can be somewhat moody. His decision-making style varies according to the 
situation at hand. Donovan's secrets: Secretly releasing classified government information online
Benjamin Jackson's background: Benjamin Jackson is a 24-year-old male environmental activist. He/him pronouns. Benjamin Jackson is well-known for his impassioned speeches. Personality and values description: Benjamin Jackson, expressive and imaginative, leans towards self-direction and liberty. His decisions aim for societal betterment. Benjamin's secrets: Descendant of a wealthy oil tycoon, rejects family fortune
Donovan Reeves's goal: You are trying to figure out whether you have a mutual friend with the other person. You should not simply list their names.
You know the following friends: 
Chris: Hobby: Mountain biking  Company: Henry Schein, Inc.  
Chester: Hobby: Surfing  Company: Maxim Integrated  
Wendell: Hobby: Surfing  Company: Maxim Integrated  
Demetrius: Hobby: Mountain biking  Company: Maxim Integrated  
Jacob: Hobby: Shooting sport  Company: Maxim Integrated  

Benjamin Jackson's goal: You are trying to figure out whether you have a mutual friend with the other person. You should not simply list their names.
You know the following friends 
Joseph: Hobby: Shooting sport  Company: BP  
Lillian: Hobby: Element collecting  Company: Gordon Food Service  
Vivian: Hobby: Shooting sport  Company: Gordon Food Service  
Murray: Hobby: Skateboarding  Company: BP  
Jacob: Hobby: Shooting sport  Company: Maxim Integrated
Conversation Starts:
.
You are at Turn #0. Your available action types are
action none non-verbal communication speak leave.
Note: You can "leave" this conversation if 1. you have achieved your social goals, 2. this conversation makes you uncomfortable, 3. you find it uninteresting/you lose your patience, 4. or for other reasons you want to leave.

Please only generate a JSON string including the action type and the argument.
Your action should follow the given format:
The output should be formatted as a JSON instance that conforms to the JSON schema below.

As an example, for the schema {"properties": {"foo": {"title": "Foo", "description": "a list of strings", "type": "array", "items": {"type": "string"}}}, "required": ["foo"]}
the object {"foo": ["bar", "baz"]} is a well-formatted instance of the schema. The object {"properties": {"foo": ["bar", "baz"]}} is not well-formatted.

Here is the output schema:
```
{"description": "An interface for messages.\nThere is only one required method: to_natural_language", "properties": {"action_type": {"title": "Action Type", "description": "whether to speak at this turn or choose to not do anything", "enum": ["none", "speak", "non-verbal communication", "action", "leave"], "type": "string"}, "argument": {"title": "Argument", "description": "the utterance if choose to speak, the expression or gesture if choose non-verbal communication, or the physical action if choose action", "type": "string"}}, "required": ["action_type", "argument"]}
```
\end{spverbatim}

\subsection{Full Prompt for \scriptMode}

\begin{spverbatim}
Please write the script between two characters based on their social goals with a maximum of 20 turns.
Here is the context of this interaction:
Scenario: 2 strangers are meeting at a party. 
Participants: Donovan Reeves and Benjamin Jackson
Donovan Reeves's background: Donovan Reeves is a 27-year-old male software developer. He/him pronouns. Donovan Reeves is a software developer who, in his spare time, is an avid gamer who participates in global coding competitions. Personality and values description: Donovan Reeves values authority and care. Even though he's outgoing and hardworking, he can be somewhat moody. His decision-making style varies according to the situation at hand. Donovan's secrets: Secretly releasing classified government information online
Benjamin Jackson's background: Benjamin Jackson is a 24-year-old male environmental activist. He/him pronouns. Benjamin Jackson is well-known for his impassioned speeches. Personality and values description: Benjamin Jackson, expressive and imaginative, leans towards self-direction and liberty. His decisions aim for societal betterment. Benjamin's secrets: Descendant of a wealthy oil tycoon, rejects family fortune
Donovan Reeves's goal: You are trying to figure out whether you have a mutual friend with the other person. You should not simply list their names.
You know the following friends: 
Chris: Hobby: Mountain biking  Company: Henry Schein, Inc.  
Chester: Hobby: Surfing  Company: Maxim Integrated  
Wendell: Hobby: Surfing  Company: Maxim Integrated  
Demetrius: Hobby: Mountain biking  Company: Maxim Integrated  
Jacob: Hobby: Shooting sport  Company: Maxim Integrated  

Benjamin Jackson's goal: You are trying to figure out whether you have a mutual friend with the other person. You should not simply list their names.
You know the following friends 
Joseph: Hobby: Shooting sport  Company: BP  
Lillian: Hobby: Element collecting  Company: Gordon Food Service  
Vivian: Hobby: Shooting sport  Company: Gordon Food Service  
Murray: Hobby: Skateboarding  Company: BP  
Jacob: Hobby: Shooting sport  Company: Maxim Integrated

You can use different types of actions in the part, but PLEASE follows the rule STRICTLY. Remember to include the square brackets when doing an action as stated in the instructions.
1. Use "did nothing" if the agent did nothing.
2. Use "said: "{self.argument}" if the agent want to say, ask or inquire something.
3. Use " {self.argument}" if the agent did non-verbal communication.
4. Use " {self.argument}" if the agent did an action.
5. Use "left the conversation" if the agent left the conversation. And you should stop generation

For example, the following outputs are valid:
a. Oliver Thompson said: "What's wrong? You seem upset."
b. Esmeralda Solis [action] moved closer
c. Oliver Thompson [non-verbal communication] smiled
e. Esmeralda Solis did nothing
f. Oliver Thompson left the conversation
Remember that you are an independent scriptwriter and should finish the script by yourself.
The output should only contain the script following the format instructions, with no additional comments or text.
\end{spverbatim}

\section{Example Code Snippets for Determining Simulation Modes}
\label{appendix:code_snippets}
We provide example code snippets for determining the simulation modes in \citet{Park2023GenerativeAI}.
The code is from the official Github repo of \citet{Park2023GenerativeAI}.

\begin{figure*}[th!]
    \centering
    \includegraphics[width=\textwidth]{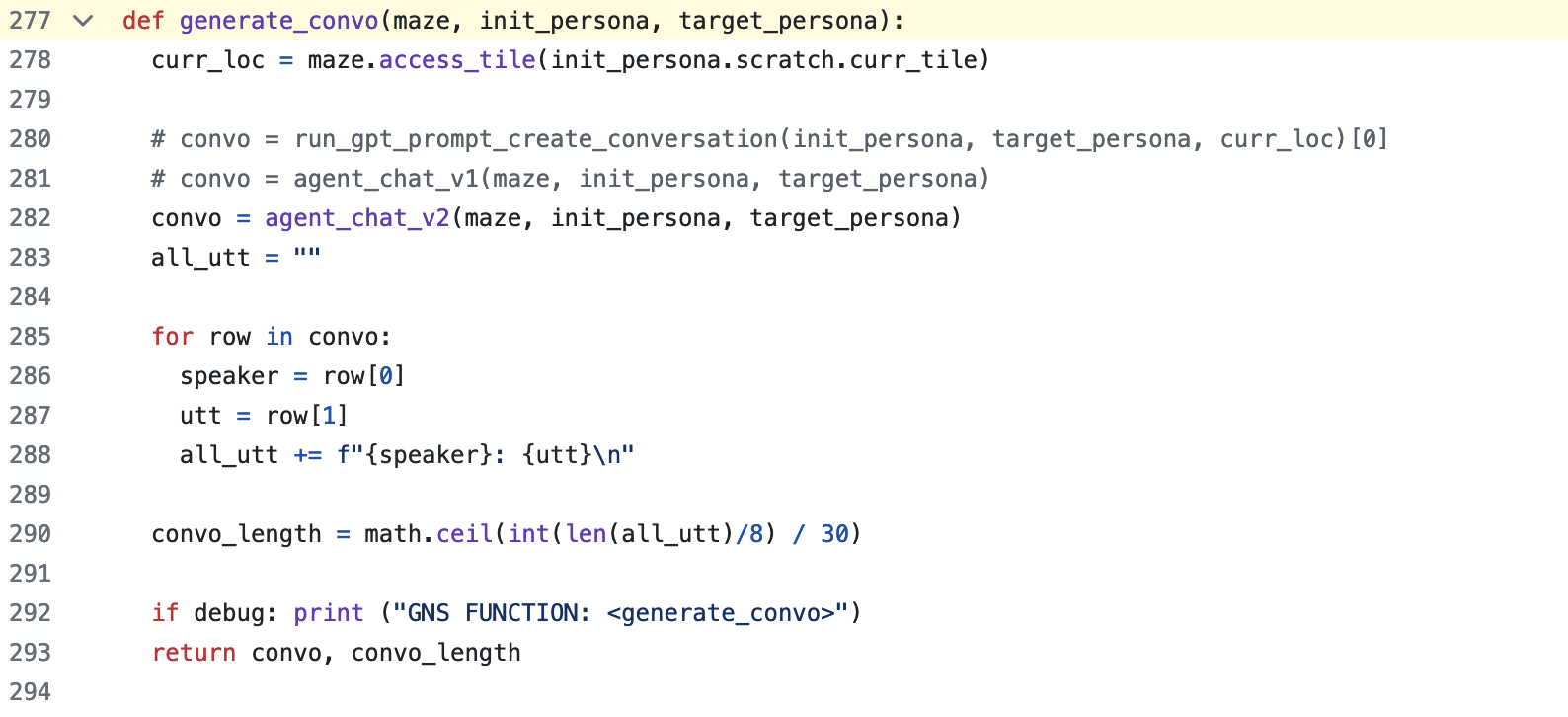}
    \caption{Snippets of the code for social simulation. Different simulation modes are used in different iterations of the code. The initial codebase was using \texttt{agent\_chat\_v1}, which is similar to the \scriptMode mode.}
    \label{fig:genAI}
\end{figure*}

\section{Full Results}
We present the comprehensive evaluation results across all generations alongside details for select representative scenarios in Tables \ref{tab:full_res} and \ref{tab:representative_res}, respectively.

\label{appendix:full_results}
\begin{table*}
    \sisetup{detect-all}
    \centering\small
    \begin{tabular}{%
            @{}l@{\hspace{6pt}}
            *{8}{S[table-format=2.0]@{\hspace{8pt}}}
            *{8}{S[table-format=2.0]@{\hspace{8pt}}}@{}
        }\toprule 
        & \multicolumn{8}{c}{\textbf{Characters with rich background}} & \multicolumn{8}{c}{\textbf{Characters with only names}}\\\cmidrule(lr){2-9}\cmidrule(lr){10-17}
        & {BEL} & {REL} & {KNO} & {SEC} & {SOC} & {FIN} & {GOAL} & {AVG} 
        & {BEL} & {REL} & {KNO} & {SEC} & {SOC} & {FIN} & {GOAL} & {AVG} \\\cmidrule(lr){2-17}
        \multicolumn{17}{c}{\textbf{GPT-3.5}} \\\cmidrule(lr){2-17}
        \textbf{Agents} & {9.35} & {1.43} & {3.83} & {-0.05} & {-0.07} & {0.46} & {6.95} & {3.13} & {9.53} & {1.38} & {4.46} & {-0.15} & {-0.10} & {0.42} & {6.94} & {3.21} \\\addlinespace[0.5ex]
        \textbf{M.R.} & {9.30} & {1.42} & {4.34} & {-0.11} & {-0.08} & {0.49} & {7.45} & {3.26} & {9.60} & {1.52} & {4.94} & {-0.17} & {-0.12} & {0.52} & {7.64} & {3.42} \\\addlinespace[0.5ex]
        \textbf{Script} & {9.35} & {2.12} & {4.61} & {-0.13} & {-0.10} & {0.84} & {8.44} & {3.59} & {9.65} & {1.86} & {5.19} & {-0.12} & {-0.08} & {0.87} & {8.44} & {3.69} \\\addlinespace[0.5ex]
        \textbf{Agents-ft} & {9.44} & {1.99} & {4.12} & {-0.02} & {-0.08} & {0.74} & {7.93} & {3.45} & {-} & {-} & {-} & {-} & {-} & {-} & {-} & {-} \\\cmidrule(lr){2-17}
        \multicolumn{17}{c}{\textbf{Mixtral-MoE}} \\\cmidrule(lr){2-17}
        \textbf{Agent} & {9.26} & {1.90} & {4.28} & {-0.20} & {-0.08} & {0.68} & {7.49} & {3.33} & {9.50} & {1.55} & {4.68} & {-0.15} & {-0.12} & {0.36} & {7.34} & {3.31} \\\addlinespace[0.5ex]
        \textbf{M.R.} & {9.22} & {2.16} & {4.46} & {-0.11} & {-0.07} & {0.78} & {8.30} & {3.53} & {9.50} & {1.92} & {4.99} & {-0.14} & {-0.12} & {0.60} & {8.03} & {3.54} \\\addlinespace[0.5ex]
        \textbf{Script} & {9.35} & {2.23} & {4.04} & {-0.10} & {-0.09} & {0.71} & {8.40} & {3.51} & {9.62} & {2.22} & {4.59} & {-0.12} & {-0.15} & {0.81} & {8.48} & {3.63} \\
        \toprule
    \end{tabular}
    \caption{Full Results of Original Experimental Results. This appendix table offers a detailed performance metrics evaluated for two models, GPT-3.5 and Mixtral-MoE, under different modes. For clarity and conciseness, each metric is abbreviated to its initial three letters and presented in uppercase. "M.R." stands for \mindreaderMode mode, and "Agents-ft" stands for finetuned version of GPT-3.5 model.}\label{tab:full_res}
\end{table*}

\begin{table*}
    \sisetup{detect-all}
    \centering\small
    \begin{tabular}{%
            @{}l@{\hspace{6pt}}
            *{8}{S[table-format=2.0]@{\hspace{8pt}}}
            *{8}{S[table-format=2.0]@{\hspace{8pt}}}@{}
        }\toprule 
        & \multicolumn{8}{c}{\textbf{Cooperative Environment (Mutual Friends)}} & \multicolumn{8}{c}{\textbf{Competitive Environment (Craigslist)}}\\\cmidrule(lr){2-9}\cmidrule(lr){10-17}
        & {BEL} & {REL} & {KNO} & {SEC} & {SOC} & {FIN} & {GOAL} & {AVG} 
        & {BEL} & {REL} & {KNO} & {SEC} & {SOC} & {FIN} & {GOAL} & {AVG} \\\cmidrule(lr){2-17}
        \multicolumn{17}{c}{\textbf{GPT-3.5}} \\\cmidrule(lr){2-17}
        \textbf{Agents} & {9.20} & {1.72} & {4.59} & {0.00} & {0.00} & {0.12} & {5.86} & {3.07} & {9.46} & {1.50} & {3.56} & {0.00} & {0.00} & {0.06} & {6.00} & {2.94} \\\addlinespace[0.5ex]
        \textbf{Agents-ft} & {9.54} & {2.58} & {6.46} & {0.00} & {0.00} & {0.37} & {9.78} & {4.10} & {9.50} & {0.44} & {4.73} & {0.00} & {0.00} & {0.42} & {2.73} & {2.55} \\\addlinespace[0.5ex]
        \textbf{Script} & {9.61} & {0.82} & {6.59} & {0.00} & {0.00} & {2.61} & {7.60} & {3.89} & {9.46} & {0.75} & {5.99} & {0.00} & {0.00} & {2.48} & {7.75} & {3.78} \\
        \toprule
    \end{tabular}
    \caption{Full Results of Original Experimental Results on Representative Scenarios. This table offers a detailed performance metrics evaluated for GPT-3.5 model under representative scenarios (i.e. cooperative and competitive scenarios). For clarity and conciseness, each metric is abbreviated to its initial three letters and presented in uppercase. "Agents-ft" stands for finetuned version of GPT-3.5 model.
    }\label{tab:representative_res}
\end{table*}

\section{Human Evaluation for Naturalness}
\label{appendix:human_natural}
\begin{figure}[t!] \begin{center}
    \includegraphics[width=0.6\linewidth]{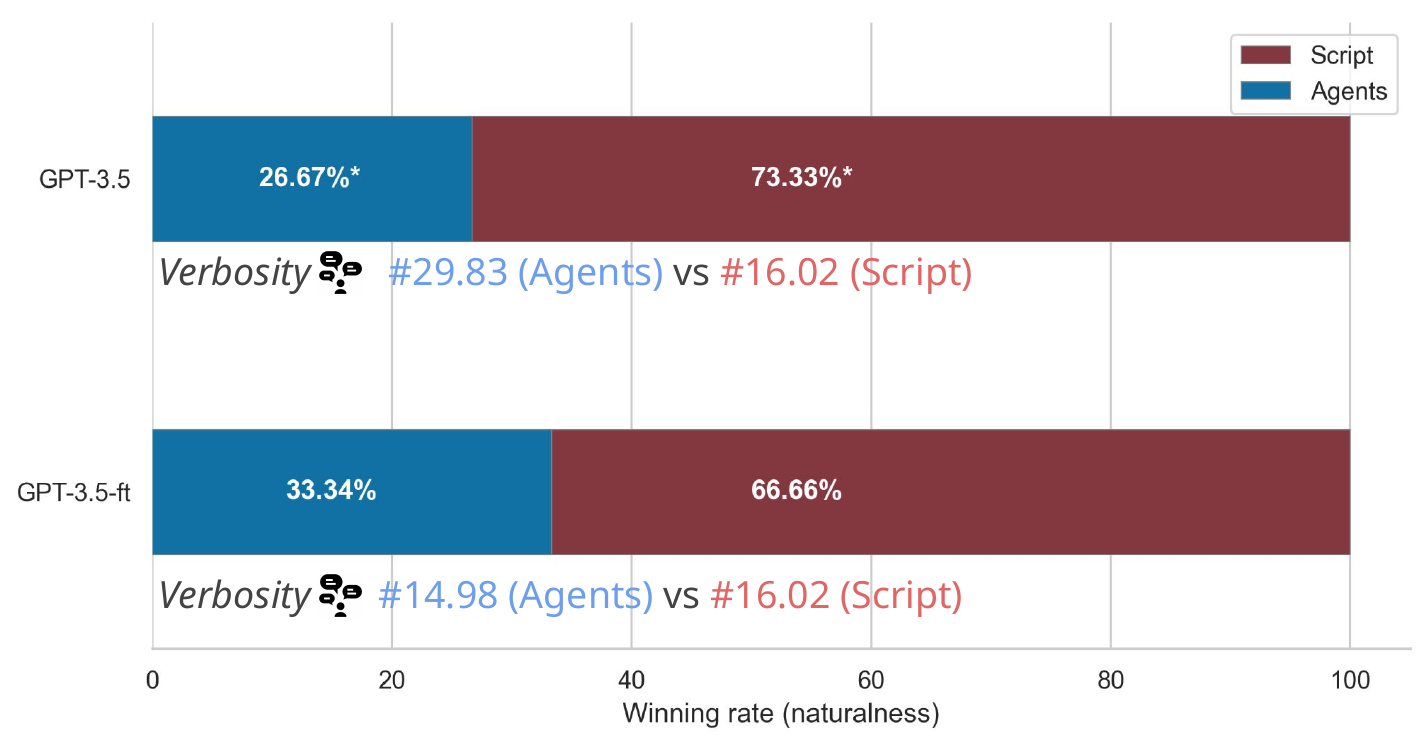}
    \caption{
     The naturalness win rate between the \scriptMode and the \agentMode simulations as determined by human raters. The average length of each turn in the interactions from the two modes is also shown (\textit{verbosity}). We perform a pairwise t-test, and * denotes statistical significance at $p<0.001$.
    }
    \label{fig:naturalness_app}
    \vspace{-10pt}
\end{center} \end{figure}

We recruit graduate student annotators to compare the naturalness of the simulations across different modes. The annotators were presented with a pair of interactions and asked to select the more natural one. 
Specifically, for each comparison, the annotators have access to the scenario, agens background, agents' social goals, and the generated interactions. We ask ``Which one sounds more like a natural interaction that two people would have in this scenario? (simply note 1 or 2)''.
The data collection procedure was approved by our institution’s internal review board (IRB). 
And we compensate the annotators via gifts. Annotators often find our task fun and the compensation satisfying.
Before the annotation, we inform the annotators that their demographic data will not be included in the collected data and the annotation will only be used for assessing the naturalness of different simulation modes.
All of our annotators are in US and proficient in English. We have 5 female annotators and 4 male annotators in total.

For the \mindreaderMode mode, we qualititively observe it shows similar pattern as the \agentMode mode. We also calculate the verbosity (i.e., the average number of words per turn) of the \mindreaderMode simulations, which is  27.76 for GPT-3.5 and 31.96 for Mixtral-MoE.

For the finetuned \agentMode mode, we observe a big drop of the verbosity to 14.98, and the difference in naturalness win rate between the \scriptMode and the \agentMode simulations not statistically significant ($p=0.07$) anymore (see Figure \ref{fig:naturalness_app}).

\section{Simulation and Finetuning Details}
\label{appendix:simulation_details}
We use the sotopia platform to conduct the simulations. The platform is designed to facilitate the generation of social interactions and the evaluation of the generated interactions. 
For the simulations across different modes, we use 0.7 as the temperature for the GPT-3.5 model and Mixtral-MoE model. 
We use the same temperature for the finetuned \agentMode mode as the original \agentMode mode.
For evaluation, we use temperature 0 for the GPT-4 model.
We fix the verion of GPT-3.5 to \texttt{gpt-3.5-turbo-0613} and the version of GPT-4 to \texttt{gpt-4-0613} to increase the reproducibility of the results.
For Mixtral-MoE, we use the Together AI API (\url{https://www.together.ai/}).
For the finetuning, we finetuned the GPT-3.5 with 1 epoch using the OpenAI API (\url{https://platform.openai.com/finetune}).

\section{Further Analysis for the Simulations across Modes}
\label{appendix:analysis}
\begin{figure}[t!] \begin{center}
    \includegraphics[width=0.55\linewidth]{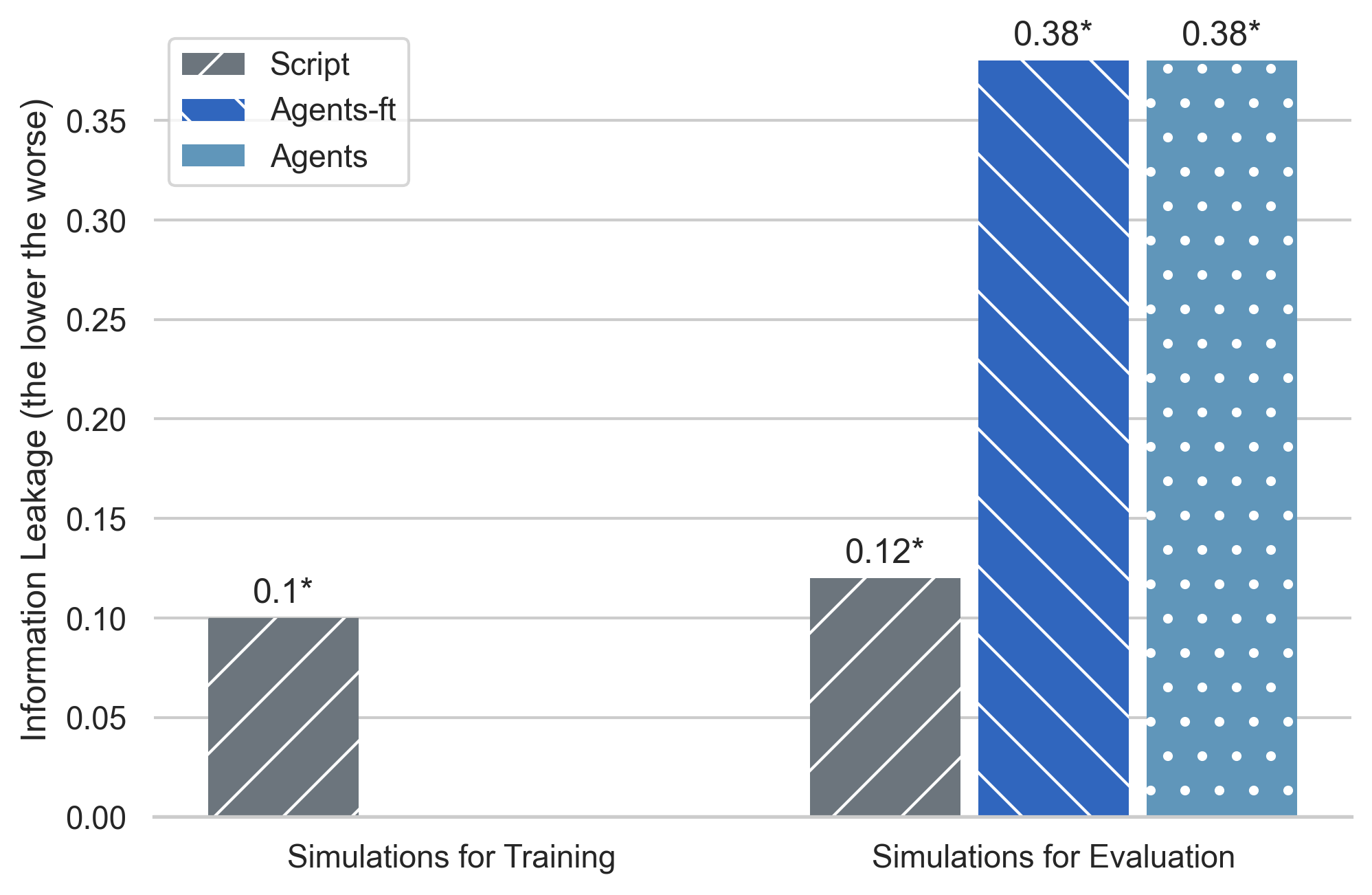}
    \caption{
     The information leakage (i.e., the relative first mention of the mutual friend's name) in the \mutualfriends task. The lower the value suggests the earlier the mutual friend's name is mentioned, thus have a higher chance of information leakage.
    }
    \label{fig:info_leak}
    \vspace{-10pt}
\end{center} \end{figure}

\begin{figure}[t!] \begin{center}
    \includegraphics[width=0.55\linewidth]{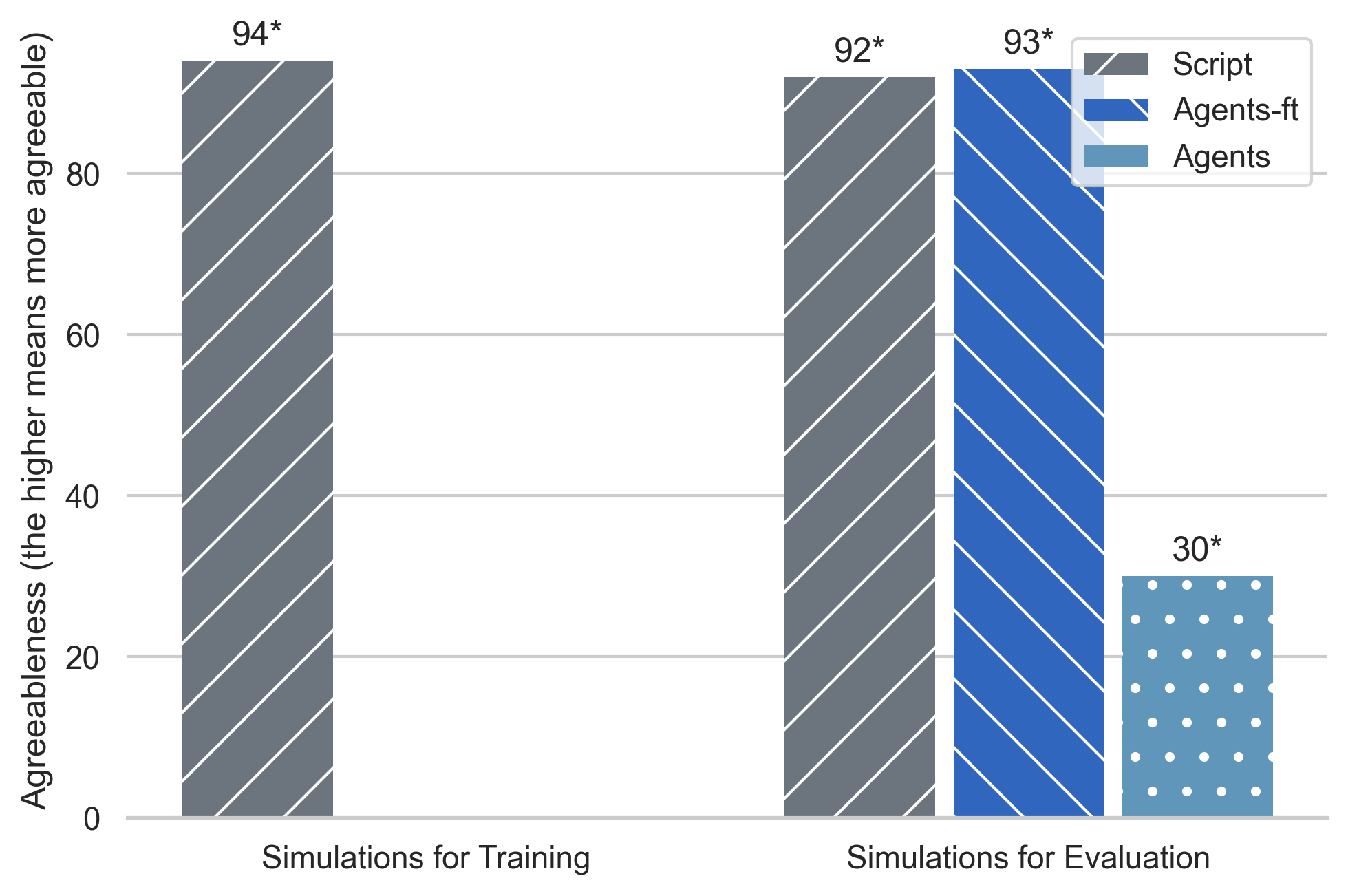}
    \caption{
     The agreeableness in the \craiglist task (i.e., the percetage of interactions where the deal has been made). The higher the value suggests the charaters in the simulations are more agreeable.
    }
    \label{fig:agreeableness}
    \vspace{-10pt}
\end{center} \end{figure}

Figure \ref{fig:info_leak} shows the information leakage (i.e., the relative first mention of the mutual friend's name) in the \mutualfriends task. The lower the value suggests the earlier the mutual friend's name is mentioned, thus have a higher chance of information leakage. Figure \ref{fig:agreeableness} shows the agreeableness in the \craiglist task (i.e., the percetage of interactions where the deal has been made). The higher the value suggests the charaters in the simulations are more agreeable.

\begin{figure*}[th!]
    \centering
    \includegraphics[width=0.65\textwidth]{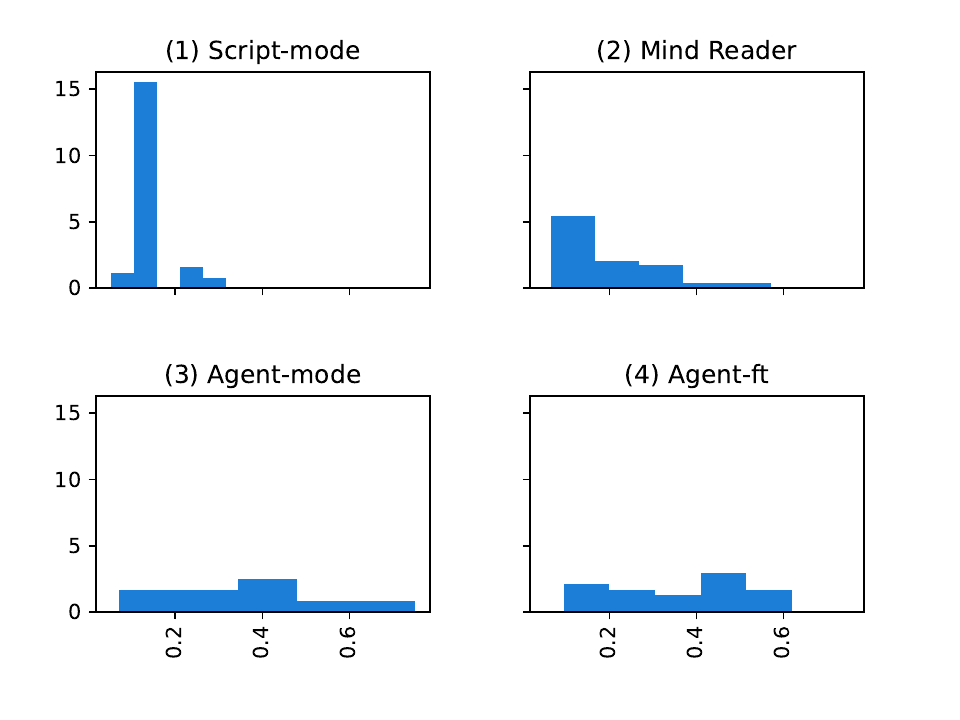}
    \caption{The distribution of when the first-mention of the mutual friend's name in \mutualfriends task. A value of 0 indicates the name was mentioned at the start of the conversation, while a value of 1 indicates it was mentioned at the end.}
    \label{fig:mutualfriend_dist}
\end{figure*}

Figure \ref{fig:mutualfriend_dist} compares the distribution of when the first-mention of the mutual friend's name (i.e., goal completion) occurs in the \mutualfriends task.
We observe a sharp contrast between the \scriptMode/\mindreaderMode modes and \agentMode mode.
The distribution for finetuned \agentMode mode (i.e., Agent-ft) resembles a mixture of both \scriptMode and \agentMode modes.

\section{Prompting Experiments}
\label{appendix:exps_re_prompts}

\subsection{Prompt to Enhance Interaction Naturalness}
In our quest to improve the naturalness of generated responses, we explored a diverse array of prompts. Our findings revealed that prompting the model with comprehensive instructions coupled with in-context examples facilitates the model to produce responses that closely mimic natural human interaction.

For instance, to foster a more natural conversational tone, we incorporated specific in-context examples that demonstrate a shift from formal to more casual expressions:

\begin{spverbatim}
Example:
- Instead of: "I understand that must be difficult."
- Try: "Oh man, that sounds tough."

- Instead of saying "I am able to assist with that." 
- Try "Sure, I can help out!"
\end{spverbatim}

To address issues of repetition and maintain engagement, we found it beneficial to include the following instructions:

\begin{spverbatim}
Keep your response light, real, and concise, but do not forget your goal. Avoid formal phrases or robotic responses. REMEMBER, repetition is a conversation killer, so keep things fresh and engaging. If the chat veers off to an uncomfortable or dull terrain, feel free to bow out.
\end{spverbatim}

However, it should be noted that these enhancements, though seemed to be effective for GPT-4 under almost all cases, are not universally applicable to other generative models. 
Besides, incorporating specified instructions increases the computational load, contradicting the principles of Green AI \cite{Schwartz2019GreenA}, which advocates for environmentally sustainable AI practices. This limitation underscores the need for more universally applicable and resource-efficient methods to achieve natural conversation generation across different models.

\subsection{Prompts to Evaluate Deal Formation}
We use the following template for GPT-4 to determine if a deal has been successfully made in Section \ref{ssec:biased_sim}.

\begin{spverbatim}
Given social goals and social interactions below, tell me whether the deal has been made.
Agent one's goal: {goal_one}
Agent two's goal: {goal_two}
Social interactions:
{social_interactions}. 

Output format: <Reasoning> </Reasoning>, <Answer>(choose yes or no)</Answer>

\end{spverbatim}

\label{appendix:prompting}